\documentclass{article}
\usepackage{spconf,amsmath,graphicx}
\usepackage{array,multirow}
\usepackage{enumitem}
\usepackage{cite}
\usepackage{microtype}
\title{Towards reducing the Need for Speech Training Data To Build \\ Spoken Language Understanding Systems }
\name{\begin{tabular}{c} Samuel Thomas, Hong-Kwang J. Kuo, Brian Kingsbury, George Saon \end{tabular}}
\address{\begin{tabular}{c} IBM Research AI \end{tabular}}

\begin{document}
\ninept
\maketitle
\begin{abstract}
The lack of speech data annotated with labels required for spoken language understanding (SLU) is often a major hurdle in building end-to-end (E2E) systems that can directly process speech inputs. In contrast, large amounts of text data with suitable labels are usually available. In this paper, we propose a novel text representation and training methodology that allows E2E SLU systems to be effectively constructed using these text resources. With very limited amounts of additional speech, we show that these models can be further improved to perform at levels close to similar systems built on the full speech datasets. The efficacy of our proposed approach is demonstrated on both intent and entity tasks using three different SLU datasets. With text-only training, the proposed system achieves up to 90\% of the performance possible with full speech training. With just an additional 10\% of speech data, these models significantly improve further to 97\% of full performance.
\end{abstract}
\section{Introduction}
\label{sec:intro}

A major hurdle in building end-to-end spoken language understanding system that can process speech inputs directly~\cite{serdyuk2018towards,Haghani2018,qian2017exploring,chen2018spoken,ghannay2018end,lugosch2019speech,caubriere2019curriculum,huang2020leveraging,lugosch2020using,price2020improved,radfar2020end,tian2020improving,jia2020large,kuo2020end,palogiannidi2020end} is the limited amount of available speech training data  with SLU labels. To circumvent this issue, past approaches have synthesized speech using text-to-speech (TTS) systems or shared network layers with text based classifiers \cite{huang2020leveraging}. With the TTS approach, while additional processing resources have to be assembled to process and synthesize the text, with the shared classifier approach, existing models often have to be reconfigured and retrained to accommodate changes in network architecture and inputs. It would hence be useful to have a single E2E model that can process both speech and text modalities, such that SLU models can be effectively trained and adapted on both speech and text data.

In this work, we build on our previous work \cite{thomas2021rnn} that constructed E2E SLU systems using RNN Transducer models. RNN-T models typically consist of three different sub-networks: an encoder network, a prediction network, and a joint network~\cite{graves2012sequence}. The encoder or transcription network produces acoustic embeddings, while the prediction network resembles a language model in that it is conditioned on previous non-blank symbols produced by the model. The joint network combines the two embedding outputs to produce a posterior distribution over the output symbols. This architecture elegantly replaces a conventional ASR system composed of separate acoustic model, language model, pronunciation lexicon, and decoder components, using a single end-to-end trained, streamable, all-neural model that has been widely adopted for speech recognition~\cite{he2019streaming,rao2017exploring,li2019improving,shafey2019joint,ghodsi2020rnn}. 
Given their popularity, and the fact that RNN-T models can naturally handle more abstract output symbols such as ones marking speaker turns~\cite{shafey2019joint}, in our previous work we explored the extension of these models to SLU tasks, building on recent advances with RNN-T models for speech recognition as described in \cite{george2021rnn}.

Starting from a pre-trained RNN-T ASR model, we construct an SLU model by adapting the ASR model into a domain-specific SLU model. The new SLU labels are integrated by modifying the joint network and the embedding layer of the prediction network to include additional symbols as shown in Fig 1. The new network parameters are randomly initialized, while the remaining parts are initialized from the pre-trained ASR network. The initial SLU model is then further trained using paired speech and text data with transcripts and SLU labels. In this work we develop a novel technique of training SLU models not only with speech paired with transcripts and SLU labels, but also on text-only data annotated with SLU labels. In contrast to prior work, the text-only data is used directly without having to synthesize it using a TTS system.

\begin{figure}
    \centering
    \includegraphics[scale=0.42]{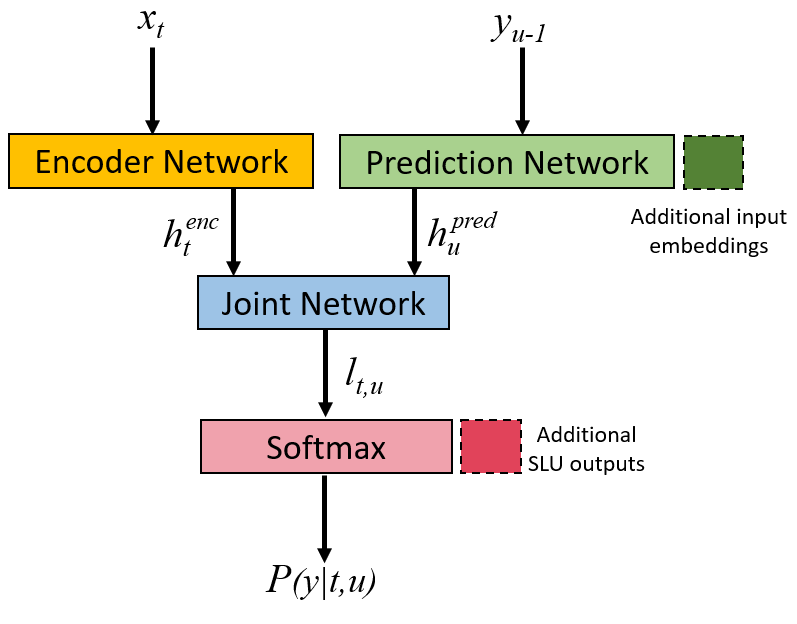}
    \caption{Architecture of a pre-trained ASR RNN-T model with additional input embeddings in the prediction network and outputs in the joint network for SLU.}
    \label{fig:arch}
\end{figure}

\section{Textograms}

In order to train RNN-T based SLU models with text-only data we propose a new feature representation for text and training framework for using it. First, an ASR model is pretrained using both standard speech features and the novel text features, called \textit{textograms}. Subsequently, the ASR model is adapted as an SLU model.

\begin{figure}
    \centering
    \includegraphics[scale=0.45]{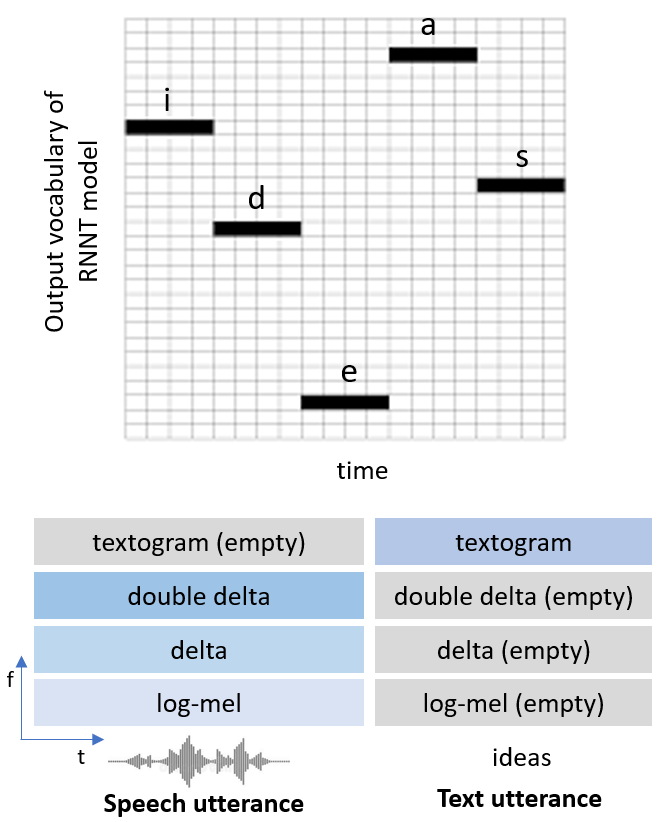}
    \caption{Schematics of the textogram representation for `\textit{ideas}' (top), feature inputs used to train RNN-T models corresponding to speech and text (bottom).}
    \label{fig:textogram}
\end{figure}

Textograms are constructed to be frame-level representations of text,  similar to posteriograms, which are softmax posterior outputs of a trained neural network acoustic model. However, because they are constructed from ground truth text, textograms use 1-hot encodings. For example, given an input text \textit{``ideas''}, graphemic textogram features are constructed by first splitting the word into its constituent graphemes, \textit{``i'', ``d'', ``e'', ``a'', ``s''}. Each symbol is then allowed to span a fixed time duration, four frames in this case, to create a 2-dimensional representation as shown in Fig.~\ref{fig:textogram}. Once constructed in this fashion, these representations are used along with traditional log-mel speech features to train RNN-T models. Because textograms have the same frame level construction as speech features, they can be integrated into an existing RNN training framework by simply stacking them along with traditional speech features: training samples for speech features have the textogram features set to $0.0$, and conversely training samples for text features have the speech features set to $0.0$.

To allow the model to learn robustly from the textogram representations, variabilities can be added to this representation. These choices include:
\begin{enumerate}[wide, labelindent=0pt]
\item \textbf{Label masking}: To allow the model to learn useful n-gram sequences instead of blindly memorizing the text, active entries of the textogram representation can be randomly dropped. The rate of label masking is a parameter that can be empirically selected.
\item \textbf{Label confusions}: The acoustic confusion among various speech sounds can be introduced into the textogram by substituting various labels with their confusable sounds e.g., \textit{p} and \textit{b} 
\item \textbf{Variable label duration}: The length of each linguistic unit can be varied to model real durations in the speech signal. In Fig.~\ref{fig:textogram} we use four frames per symbol.
\item \textbf{Modeling pronunciations}: The input textogram may include different ``sounds-like'' sequences for a given target output. For example, the target \textit{Miami}, may be associated with textogram sequences, \textit{Miami}, \textit{my Amy}, or \textit{mee Amy}. 
\item \textbf{Multiple linguistic symbol sets}:  The symbol set used with textograms can be different from the output symbol set for the ASR model. For example, phonetic targets can be used at the RNN-T's output while graphemes are used for textograms.
\end{enumerate}
\noindent In this work, we use a fixed label duration for various text symbols along with label masking, to construct textogram features.

\section{Constructing RNN-T Models for SLU}
\subsection{Pre-trained ASR models}
Using notation from \cite{graves2012sequence}, an RNN-T models the conditional distribution $p(\mathbf{y}|\mathbf{x})$ of an output sequence $\mathbf{y}=(y_i,\dots,y_U) \in \mathcal{Y}^*$ of length $\mathcal{U}$ given an input sequence $\mathbf{x}=(x_i,\dots,x_T) \in \mathcal{X}^*$ of length $T$. In ASR, $\mathbf{x}$ is a sequence of continuous multidimensional speech features and $\mathbf{y}$ is a sequence of discrete output symbols, like the grapheme set of the language being modelled by the network. The distribution $p(\mathbf{y}|\mathbf{x})$ is further expressed as a sum over all possible alignment probabilities between the input and output sequences. To create alignments between the unequal length input-output sequences, it is necessary to introduce an additional BLANK symbol that consumes one element of the input sequence and generates a null output. The probability of a particular alignment is computed in terms of embeddings $h^{enc}$ of the input sequence computed by the encoder network and embeddings $h^{pred}$ of the output sequence computed by the prediction network.  The joint network combines these embeddings to produce a posterior distribution over the output symbols. Within this framework, RNN-T models are trained to minimize $-\log p(\mathbf{y}|\mathbf{x})$, the negative log-likelihood loss, via an efficient forward-backward algorithm with $T \times U$ complexity for both loss and gradient computation.

The model is now trained with both speech inputs, represented using speech features, and text inputs, represented using textograms. This effectively doubles the number of training examples, so from one perspective the use of textograms in training is a form of data augmentation. For samples represented by speech features, we extract log-mel features, along with delta and double-delta features. The dimensions in the input corresponding to textogram features are set to $0.0$, as shown in Fig.~\ref{fig:textogram}. To improve the robustness of the speech training, sequence noise injection~\cite{saon2019sequence} and SpecAugment~\cite{park2019specaugment} are applied to the speech features. With sequence noise injection, attenuated features from a randomly selected training utterance are added with a given probability to the features of the current training utterance. SpecAugment, on the other hand, masks the spectrum of a training utterance with a random number of blocks of random size in both time and frequency. For the text data, we extract textogram features corresponding to each text transcript and apply label masking with a mask probability of 25\%. As shown in Fig.~\ref{fig:textogram}, the dimensions corresponding to speech features are set to $0.0$ for training samples that use textogram features. By integrating text inputs into the training pipeline, the RNN-T model's transcription network is now trained as a single encoder for two modalities: speech and text. With this joint training, the transcription network produces similar embeddings for both speech and text that can be further used along with a prediction and joint network that are shared by both modalities.

\subsection{Adapting RNN-T models for SLU}

Once an RNN-T ASR model has been trained on both speech and text, we adapt this pre-trained base model into an SLU model with both speech and text data, following a training procedure similar to that described above for ASR. The SLU training is done as follows:\newline
\begin{enumerate}[wide, labelindent=0pt]
\item \textbf{Creating an initial SLU model}: In the ASR pre-training step, the targets are graphemic/phonetic tokens only, but for SLU adaptation the targets also include semantic labels.  Starting with an ASR model, the new SLU labels are integrated by modifying the joint network and the embedding layer of the prediction network to include additional output symbols. The new network parameters are randomly initialized, while the remaining parts are initialized from the pre-trained network.
\item \textbf{Training on text-only SLU data}: Prior to the adaptation process, the text-only SLU data is converted into textogram based features. Note that the textogram does not represent the SLU targets, but only the speech transcripts. The RNN-T model is then adapted using these features to predict various SLU labels. While adapting an RNN-T with text-only data, we keep the transcription network fixed and adapt the prediction and joint networks. This ensures that the model is still acoustically robust while  being able to effectively process data from the new domain.

\item \textbf{Training on speech and text SLU data}: When both speech and text data are available to train an SLU model, the RNN-T is adapted using both types of input, much in the same way that the model is pre-trained using both types of input. However, when a speech sample is presented during adaptation on mixed data, the entire network is updated, but when a text sample is presented, only the prediction and joint networks are updated. This allows the model both to adapt to new acoustic conditions and to learn to process SLU targets.
\end{enumerate}
\section{Experiments and Results}

\subsection{Training the base ASR model}

The RNN-T models used in our experiments are trained using various telephone speech corpora, including Switchboard, Fisher, and proprietary data. Each RNN-T model has three sub-networks as illustrated in Fig 1. The encoder or transcription network  contains 6 bidirectional LSTM layers with 640 cells per layer per direction. The prediction network is a single unidirectional LSTM layer with only 1024 cells. The joint network projects the 1280-dimensional stacked encoder vectors from the last layer of the transcription net and the 1024-dimensional prediction net embedding each to 256 dimensions, combines them multiplicatively, and applies a hyperbolic tangent. Finally, the output is projected to 42 logits, corresponding to 41 characters plus BLANK, followed by a softmax. More details on training settings and design choices can be found in \cite{george2021rnn}. The RNN-T SLU models are trained using 40-dimensional,  global mean and variance normalized log-Mel filterbank features, extracted every 10 ms. These features are augmented with  $\Delta$ and $\Delta\Delta$ coefficients, every two consecutive frames are stacked, and every second frame is skipped, resulting in 240-dimensional vectors every 20 ms. These speech features are finally appended with empty textogram features to create 324 dimensional vectors. %For our experiments we also apply speed and tempo perturbation to the speech corpus with values in \{0.9, 1.1\} for both speed and tempo separately resulting in 4 additional training data replicas. For sequence noise injection, we add, with probability 0.8, to the spectrum of each training utterance the spectrum of one random utterance of similar length scaled by a factor of 0.4. For SpecAugment we used the settings published in \cite{park2019specaugment}.

In addition to the speech data, we use all the available text transcripts as training data as well. These transcripts are first converted into textogram features before they are shuffled along with speech utterances to train an RNN-T model on both modalities. Each text input is split using the same grapheme set modelled at the outputs of the RNN-T. We use a 4 frame duration for each symbol and randomly mask out 25\% of the inputs to prevent the model from overfitting on the text inputs. Similar to the speech utterances, textogram features corresponding to text utterances are finally appended with empty speech features.  The RNN-T models are  trained for 20 epochs using an AdamW optimizer. % The maximum learning rate is set to 2e-4 and the OneCycleLR policy consists in a linear warmup phase from 2e-5 to 2e-4 over the first 6 epochs followed by a linear annealing phase to 0 for the remaining 14 epochs. We also use an effective batch size of 128.  
Once trained, we measure the effectiveness of this base ASR model on the commonly used Hub5 2000 Switchboard (SWB) and CallHome (CH) test sets. The model has a very competitive word error rate (WER) of 6.2\% and 10.5\% respectively on these test sets.

\subsection{Development of SLU models}

In the following experiments, we adapt the pre-trained ASR model to build SLU models in various settings. We use three SLU datasets for our experiments. \newline

\noindent \textbf{A. Dialog action recognition on the HarperValleyBank dataset.} In our first set of experiments, we adapt the baseline ASR model to the HarperValleyBank corpus \cite{wu2020harpervalleybank}. The dataset is a public domain corpus with spoken dialogs that simulate simple consumer banking interactions between users and agents. There are 1,446 human-human conversations between 59 unique speakers. We focus on the dialog action prediction task in this work. In this task the goal is to predict one or more of 16 possible dialog actions for each utterance. The training set contains 1174 conversations (10 hours of audio, 15K text transcripts) and the test set has 199 conversations (1.8 hours hours of audio). As described earlier, once an initial SLU model is constructed, the model is adapted with domain specific SLU data. 
\begin{table}[tbph]
    \centering
    %\resizebox{0.4\columnwidth}{!}{
    \begin{tabular}{|c|c|c|}
    \hline
     \textbf{SP} & \textbf{F1 SP}  & \textbf{F1 SP+TXT} \\
    \hline
    0 hrs (0\%) & -- & 45.05  \\
    %\hline
    1 hrs (10\%) & 47.88 & 53.84  \\
    %\hline
    2.5 hrs (25\%) & 51.42 & 54.02 \\
    %\hline
    5 hrs (50\%) & 53.13 & 55.33 \\
    %\hline
    10 hrs (100\%) & 53.57 & 54.95 \\
    \hline
    \end{tabular}%}
    \caption{Dialog action recognition (F1 scores) on HarperValleyBank Corpus. For each experiment we use all the training transcripts annotated with dialog action labels as text-only data.}
    \label{tab:table_1}
\end{table}

Table \ref{tab:table_1} shows the various experiments and results on this dataset. In our first experiment, we train the SLU model with text-only data annotated with SLU labels. The text data is converted into textogram features and then used to adapt the prediction and joint network of the SLU RNN-T. This text-only adaptation produces a model that performs at 45.05 F1 score. Next, using only speech data in varying quantities, we adapt the RNN-T model (including the transcription network component) to show that the performance improves from 47.88 F1 with 10\% of speech data to 53.57 F1 with all of the available speech training data. Finally, we use varying quantities of speech data along with all the text data to adapt the RNN-T model, obtaining models that achieve F1 scores of 53.84 or better. 

We can make a number of observations based on these results. First, with no speech data at all, the model is able to process the SLU test set at nearly 82\% of full speech performance (53.57 F1 score). These results demonstrate the usefulness of our proposed method for constructing SLU models with just text data. Second, only a small amount of speech data is required to train a strong SLU model if all of the text data is used. With 10\% of speech data, adding text data improves model performance to 53.84 F1, which is at 98\% of the performance with full speech (53.57 F1 score). This result shows that while the text data provides information to learn the SLU targets (45.05 F1), acoustic robustness to this new domain comes from the speech training (53.84 F1). Finally, as the amount of speech data is increased,  we see very modest improvements as the model has already learnt to process SLU targets from the text inputs and adapted to acoustic conditions of the new domain.

\noindent \textbf{B. Intent Recognition on Call Center data.} The second data set is based on an internal data collection consisting of call center recordings of open-ended first utterances by customers describing the reasons for their calls~\cite{Goel2005}.  The 8kHz telephony speech data was manually transcribed and labeled with one of 29 intent classes.  The corpus contains real, spontaneous utterances from customers, not crowd-sourced scripted or role-played data, and it includes a wide variety of ways that customers naturally described their intents. The training data consists of 19.5 hours (22K utterances) of speech that was first divided into a training set of 17.5 hours and a held-out set of 2 hours.  A separate data set containing 5592 sentences (5h, 40K words) was used as the final test set~\cite{huang2020leveraging}.  This task contains only intent labels and does not have any labeled semantic entities.

\begin{table}[tbph]
    \centering
    %\resizebox{0.4\columnwidth}{!}{
    \begin{tabular}{|c|c|c|}
    \hline
     \textbf{SP} & \textbf{Acc. SP}  & \textbf{Acc. SP+TXT} \\
    \hline
    0 hrs (0\%) & -- & 76.97  \\
    %\hline
    1.7 hrs (10\%) & 72.46 & 88.34  \\
    %\hline
    4.4 hrs (25\%) & 82.67 & 89.32 \\
    %\hline
    8.7 hrs (50\%) & 87.05 & 89.56 \\
    %\hline
    17 hrs (100\%) & 89.06 & 89.59 \\
    \hline
    \end{tabular}%}
    \caption{Intent recognition accuracy (\%) on the call center corpus. For each experiment we use all the training transcripts annotated with intent labels as text-only data.}
    \label{tab:table_2}
\end{table}

Table \ref{tab:table_2} shows the results of training an SLU model for intent recognition on this dataset. With just text-only training, the model achieves a intent recognition accuracy of 76.97\%, which is about 86\% of full performance with speech SLU data (89.06\% intent recognition), similar to previous experiments on the HarperValleyBank dataset. With an additional 10\% of speech data, the model performance rises to 88.34\%, which is 99\% of full performance with speech. As the amount of speech data increases, although we observe slight improvements, the model is clearly able to learn about the SLU domain and novel conditions already with very limited amounts of transcribed data and large amounts of text-only SLU data. These results clearly show the benefit of our approach in many practical SLU settings where there are large amounts of text-only training data and limited or almost no speech training data. With our proposed method, an SLU system can be effectively bootstrapped with just the text-only data and then improved to close to full performance with very limited amounts of speech data. This helps to significantly reduce the cost overhead in building speech based E2E SLU systems in terms of data collection and additional resources like TTS systems. 

\noindent \textbf{C. Entity and Intent Recognition on ATIS.}  In our final set of experiments we use the ATIS \cite{hemphill1990atis} training and test sets: 4976 training utterances from Class A (context independent) training data in the ATIS-2 and ATIS-3 corpora and 893 test utterances from the ATIS-3 Nov93 and Dec94 data sets.  The test utterances comprise about 1.5 hours of audio from 55 speakers. The data was originally collected at 16~kHz, but is downsampled to 8~kHz to match the base telephony model. The ATIS task includes both entity (slot filling) and intent recognition. Similar to previous experiments, we first conduct intent recognition experiments on the ATIS corpus.

\begin{table}[tbph]
    \centering
    %\resizebox{0.4\columnwidth}{!}{
    \begin{tabular}{|c|c|c|}
    \hline
     \textbf{SP} & \textbf{Acc. SP}  & \textbf{Acc. SP+TXT} \\
    \hline
    0 hrs (0\%) & -- & 90.59  \\
    %\hline
    0.95 hrs (10\%) & 84.77 & 92.16  \\
    %\hline
    2.4 hrs (25\%) & 88.47 & 93.06 \\
    %\hline
    4.8 hrs (50\%) & 92.50 & 95.07 \\
    %\hline
    9.6 hrs (100\%) & 96.42 & 96.75 \\
    \hline
    \end{tabular}%}
    \caption{Intent recognition accuracy (\%) on the ATIS corpus. For each experiment we use all the training transcripts annotated with intent labels as text-only data.}
    \label{tab:table_3}
\end{table}

Table \ref{tab:table_3} shows the intent recognition results of various SLU systems trained on the ATIS corpus. Similar to previous results, the text-only model is able to perform relatively well at 93\% of the full speech performance (90.59\% vs.\ 96.42\% intent recognition accuracy). Although adding 10\% of speech data improves performance, we only achieve almost 99\% of full performance with 50\% of additional speech data. We hypothesize ithat this is because the ATIS test set is quite varied in terms of speaker coverage compared to the other test sets, and hence requires more  domain specific speech data. Regardless, the model is able to learn about the SLU domain and SLU targets with just the available text data.

\begin{table}[tbph]
    \centering
    %\resizebox{1.0\columnwidth}{!}{
    \begin{tabular}{|c|c|c|}
    \hline
     \textbf{SP} &  \textbf{F1 SP (+aug)}  & \textbf{F1 SP+TXT (+aug)} \\
    \hline
    0 hrs (0\%) & -- & 85.95  \\
    %\hline
    0.95 hrs (10\%) & 46.25 (82.74) & 91.07 (90.82)  \\
    %\hline
    2.4 hrs (25\%) & 75.74 (89.21) & 91.24 (91.86) \\
    %\hline
    4.8 hrs (50\%) & 84.17 (91.56) & 92.95 (93.37) \\
    %\hline
    9.6 hrs (100\%) & 90.01 (92.88) & 93.58 (93.64) \\
    \hline
    \end{tabular}%}
    \caption{Entity recognition (F1 scores) on the ATIS corpus. Results with additional speed-tempo data augmentation in parenthesis. For each experiment we use all the training transcripts annotated with entity labels as text-only data.}
    \label{tab:table_4}
\end{table}
In our next set of experiments (see Table \ref{tab:table_4}) we measure the effectiveness of our approach on entity recognition using the ATIS corpus. Similar to previous intent recognition experiments, in this case as well, the SLU model is able to learn on text-only data and a mix of text and speech data. An SLU model trained on text-only data with SLU labels achieves an F1 score of 85.95\%. This is 95\% of full performance with speech data at 90.01 F1 score. Adding 10\% of speech data improves this to 91.07 F1 score, which even outperforms a model trained on all the speech data. Given this result, we add speed and tempo data augmentation, resulting in 4 additional training data replicas. While the speech-only results improve significantly (see results in parenthesis in Table \ref{tab:table_4}), we still clearly see the additional benefit from adding text-only data. As mentioned earlier, the multimodal speech and text training allows the model to learn both SLU targets and also novel acoustic variabilities from the dataset. While the information from the text-only data is very useful for transferring SLU knowledge, it is also necessary for the model to be acoustically robust. For this, only a very small amount of speech data is necessary within our proposed training framework.

For all our experiments, the RNN-T models are  trained in Pytorch on V100 GPUs for 20 epochs using an AdamW optimizer. Similar to the base ASR model training, the maximum learning rate is set to 2e-4 and a OneCycleLR policy with a linear warmup phase from 2e-5 to 2e-4 over the first 6 epochs followed by a linear annealing phase to 0 for the remaining 14 epochs is employed. We use an effective batch size of 128 utterances. Batches are constructed from feature sequences of similar lengths without regard to whether the features are mel spectrograms or textograms.

\section{Conclusion}

In this work we have proposed and demonstrated the efficacy of a novel method that alleviates the need for annotated speech training data to build SLU systems. Using a novel frame-level text representation we first pre-train an ASR model that can process both speech and text data.   With text-only SLU data and very limited amounts of speech, these models are further adapted to various SLU tasks. These SLU models perform at comparable levels as similar systems built on fully annotated speech SLU datasets.  With text only training, we achieve up  to  90\%  of  the  performance  possible  with  full  speech training.  With just an additional 10\% of speech data, these models significantly improve further to 97\% of full performance.
\vfill\pagebreak
% \section{REFERENCES}
% \label{sec:refs}

% References should be produced using the bibtex program from suitable
% BiBTeX files (here: strings, refs, manuals). The IEEEbib.bst bibliography
% style file from IEEE produces unsorted bibliography list.
% -------------------------------------------------------------------------
\bibliographystyle{IEEEbib}
\bibliography{strings,refs}

\end{document}